\documentclass[sigconf]{acmart}
\makeatletter                   
\def\mdseries@tt{m}             
\makeatother                    
\usepackage[plain]{fancyref}
\usepackage[draft=true]{minted} 
\usepackage{color}
\usepackage{hyperref}           
\hypersetup{
    colorlinks=true,
    linkcolor=blue,
    filecolor=red,      
    urlcolor=magenta,
    breaklinks=true,            
}
\usepackage{breakurl}           

\newcommand{\eventformat}[0]{AD{$\Delta$}ER}

\newcommand{\dtm}[0]{{$\Delta t_{max}$}}

\usepackage[capitalize]{cleveref}
\usepackage{caption}
\usepackage{subcaption}
\usepackage{float}
\usepackage[htt]{hyphenat} 
\usepackage{array}
\usepackage{arydshln}

\usepackage{svg}
\usepackage{algorithm}
\usepackage{algpseudocode}
\usepackage{pifont}
\newcommand{\cmark}{\ding{51}}%

\usepackage{multirow}

\crefname{section}{Sec.}{Secs.}
\Crefname{section}{Section}{Sections}
\Crefname{table}{Table}{Tables}
\crefname{table}{Tab.}{Tabs.}

\usepackage{hyperref}
\hypersetup{
    colorlinks=true,
    linkcolor=blue,
    filecolor=magenta,      
    urlcolor=cyan,
    pdftitle={Overleaf Example},
    pdfpagemode=FullScreen,
    }
\usepackage{graphicx}

\AtBeginDocument{%
  }

\setcopyright{rightsretained}
\acmISBN{ }
\acmDOI{}
\acmConference{}

\begin{document}
\sloppy                         


\title{An Asynchronous Intensity Representation for Framed and Event Video Sources}

\author{Andrew C. Freeman}
\email{acfreeman@cs.unc.edu}
\orcid{0000-0002-7927-8245}
\affiliation{%
  \institution{University of North Carolina}
  \city{Chapel Hill}
  \state{North Carolina}
  \country{USA}
}

\author{Montek Singh}
\email{montek@cs.unc.edu}
\affiliation{%
  \institution{University of North Carolina}
  \city{Chapel Hill}
  \state{North Carolina}
  \country{USA}
}

\author{Ketan Mayer-Patel}
\email{kmp@cs.unc.edu}
\affiliation{%
  \institution{University of North Carolina}
  \city{Chapel Hill}
  \state{North Carolina}
  \country{USA}
}








\renewcommand{\shortauthors}{Freeman et al.}

\begin{abstract}
 Neuromorphic ``event'' cameras, designed to mimic the human vision system with asynchronous sensing, unlock a new realm of high-speed and high dynamic range applications. However, researchers often either revert to a framed representation of event data for applications, or build bespoke applications for a particular camera's event data type. To usher in the next era of video systems, accommodate new event camera designs, and explore the benefits to asynchronous video in classical applications, we argue that there is a need for an asynchronous, source-agnostic video representation. In this paper, we introduce a novel, asynchronous intensity representation for both framed and non-framed data sources. We show that our representation can increase intensity precision and greatly reduce the number of samples per pixel compared to grid-based representations. With framed sources, we demonstrate that by permitting a small amount of loss through the temporal averaging of similar pixel values, we can reduce our representational sample rate by more than half, while incurring a drop in VMAF quality score of only 4.5. We also demonstrate lower  latency than the state-of-the-art method for fusing and transcoding framed and event camera data to an intensity representation, while maintaining $2000\times$ the temporal resolution. We argue that our method provides the computational efficiency and temporal granularity necessary to build real-time intensity-based applications for event cameras.
 
\end{abstract}

\maketitle

\section{Introduction}
In this paper, we introduce a method for transcoding both framed video and modern event camera data to a novel asynchronous video representation. Our representation enables rate-distortion tradeoffs to be systematically explored based on the temporal stability of individual pixels. The approach is computationally efficient, lossy compressible, and suitable for real-time vision applications.

For \textbf{framed video} representations, our method aims to address the following weaknesses. Chiefly, the sample rate is fixed and uniform across the frame, meaning that a spatial region of a video which is dynamic (changing greatly) has the same number of intensity samples as a spatial region which is relatively static (unchanging). Thus, there is much data redundancy in the raw representation, which is currently only addressed at the level of the compressor. Secondly, the precision of a framed video is bounded by the maximum representable value in a single image frame, typically 8-12 bits per color channel. There is no scheme, however, for representing temporally stable intensities spanning multiple frames of time with greater precision, relative to pixels whose intensities are temporally dynamic. We argue that framed video, if transformed to an asynchronous representation, will see improvements in both precision and temporal rate control.

In addition, we address the representational weaknesses of \textbf{event sensors}. These sensors
do not record traditional image frames; rather, each pixel independently records a sequence of \textit{events}, where a bespoke time-based event triggers when the log intensity of the pixel increases or decreases by a certain amount. This technique allows for extreme temporal precision (on the order of microseconds) and high dynamic range, but prevents the sensor from capturing data where the intensity is unchanging or slowly changing. Thus, many users of these cameras augment event sensors with frame-based image capture to fill in the informational gaps (e.g., DAVIS cameras \cite{survey,davis_dataset}). However, these framed images tend to exhibit motion blur, and they are typically processed separately from the event data due to the slow computational performance of fusing the information across modalities. Additionally, each event expresses the intensity change \textit{relative to} a previous event, so this raw representation does \textit{not} support lossy compression or expose a rate-distortion tradeoff. These sensors are rapidly gaining traction for vision tasks and high-speed robotics, but their extraordinary data rates (up to millions of events per second) and ever-increasing spatial resolutions (now reaching 1080p) are not adequately addressed from a systems perspective. Instead, most research applications merely convert event data to a number of frame-based representations. These representations severely reduce the temporal resolution of the data, diminishing a major benefit to event sensing.

We present the following analogy. Imagine that traditional cameras from different companies all had their own proprietary raw data formats, with no common representation between them. Now imagine that many video applications could only operate on raw data from the cameras of a particular company, and that all other video applications could operate only on PNG image sequences derived from cameras' raw representations. That is effectively how the event camera ecosystem looks today: there is no codec equivalent to H.26X \cite{h266} for event video, meaning that event camera systems are not employing any significant lossy compression and are hand-built for particular cameras' raw data formats. We argue that to build the most effective real-time event video systems, the application-level video representation should \textit{also} be asynchronous, and it should enable rate-driven lossy compression.

We propose a unified video representation whose fundamental data type is an asynchronous \textbf{intensity event}. Our events do not represent intensity \textit{change}, as opposed to the representations for event sensors; rather each event \textit{directly} encodes an intensity value. We can thus transcode both framed and event video data to our single representation, allowing us to address the above weaknesses with these data sources. We summarize our contributions as follows.

\begin{figure*}
     \centering
     \begin{subfigure}[t]{0.30\textwidth}
         \centering
         \includegraphics[width=\textwidth]{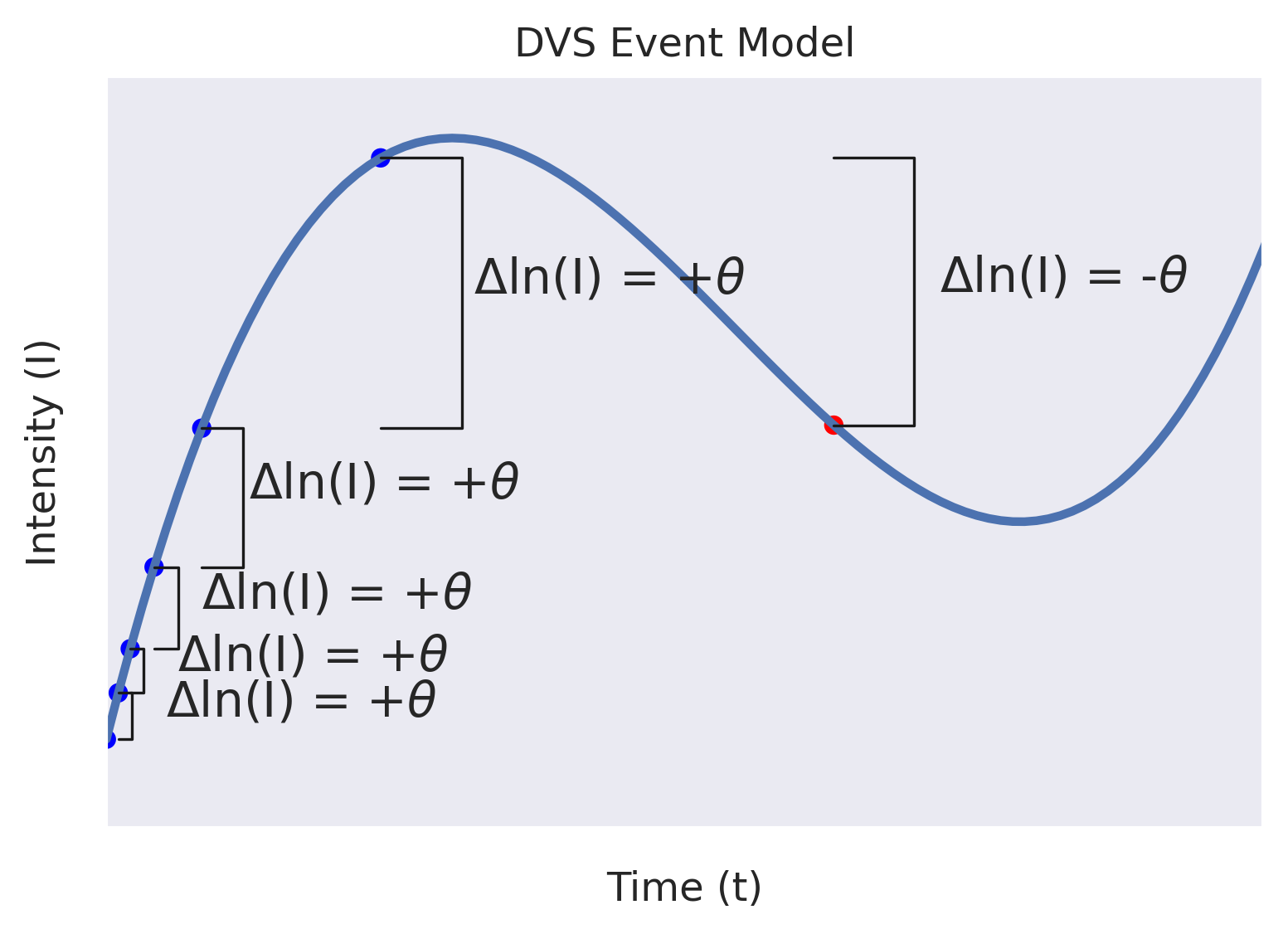}
         \caption{}
         \label{fig:dvs}
     \end{subfigure}
     \hfill
     \begin{subfigure}[t]{0.30\textwidth}
         \centering
         \includegraphics[width=\textwidth]{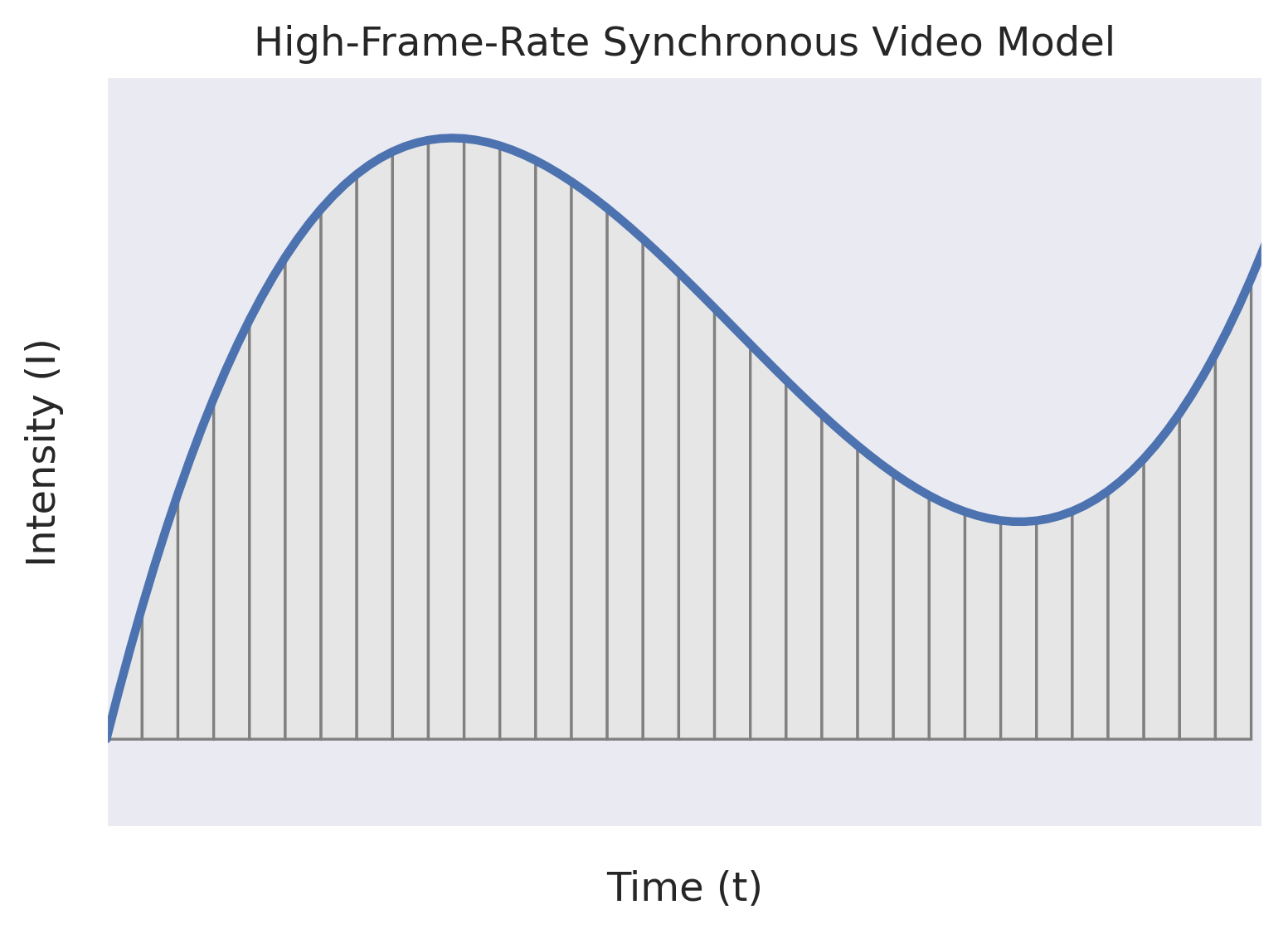}
         \caption{}
         \label{fig:framed_diagram}
     \end{subfigure}
     \hfill
     \begin{subfigure}[t]{0.30\textwidth}
         \centering
         \includegraphics[width=\textwidth]{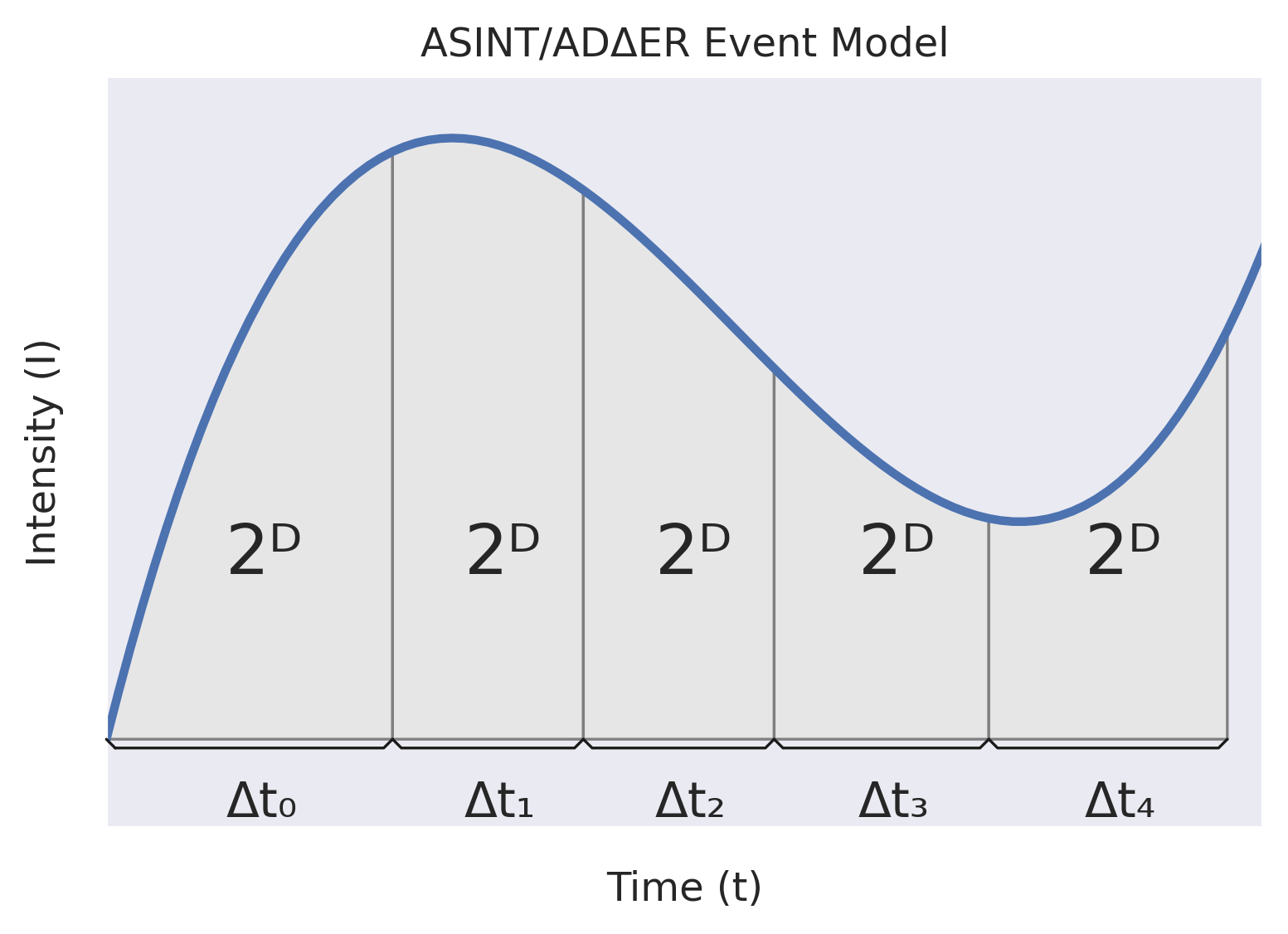}
         \caption{}
         \label{fig:asint}
     \end{subfigure}
     \caption{Comparison of video models. (a) A new DVS event is fired whenever ln(I) changes by the threshold $\theta$.  Since differences are sensed, intensity reconstruction is challenging. (b) A framed video contains one sample for each pixel at synchronous points in time, with fixed lengths between samples. (c) An \eventformat{} event indicates that the accumulated intensity units reach $2^D$.  $D$ controls the per-pixel sensitivity, which we show as constant here for illustrative purposes. In our \eventformat{} transcoder, however, $D$ adapts to changes in the intensities being integrated, to reduce the event rate.}
     \label{fig:sensortypes}
\end{figure*}

\begin{itemize}
    \item Introduce a novel asynchronous video representation and transcoder for both framed and non-framed data sources.
    \item Evaluate the speed and visual quality of our representation on framed color video sources, demonstrating extremely high precision (up to 14 bits, as opposed to 8 bits for the source videos) and a predictable rate-distortion curve. By permitting a small amount of loss, we can reduce the median event rate of our representation by 54.3\% with a median drop in VMAF quality of only 4.5.
    \item Evaluate the effect of various transcoder parameters on rate and quality.
    \item Evaluate the speed and rate properties of our representation on event camera sources with and without additional framed data, and show the qualitative differences in appearance at various rates. When targeting a temporal precision of 500 samples per second or higher, our method incurs less latency than the state-of-the-art method for frame/event fusion. In fact, our method maintains \textit{microsecond} temporal resolution for intensities that are changing. We also release the first fully open-source implementation of this state-of-the-art fusion method, with numerous improvements and optimizations for high performance.
    \item Release our open-source transcoder framework, including utilities for visual playback of our video representation.
\end{itemize}

\section{Related Work}
\subsection{Framed Video}
    
    Under the framed video capture paradigm, the  camera exposes its entire imaging plane for the same amount of time, producing a series of discrete 2D images, as in \cref{fig:framed_diagram}. Although framed video representations benefit from decades of compression research for a variety of storage needs and streaming applications, they have many limitations. 
    
    For one, dynamic range is inseparable from the source's capture sensitivity, providing a practical upper limit to the intensity range of the representation \cite{dynamic_range}. 
    Often, one records video at a high bit depth (up to 12 bits), then masters the video for viewing on both SDR and HDR displays \cite{tone_mapping}. The mastering process often involves adjusting brightness levels to show desired levels of detail. In this case, a higher-precision representation allows a filmmaker to have more flexibility in adjusting the dark and bright regions of the image during post-processing. Additionally, high precision helps to diminish the apparent edges within regions of intensity gradient.
    
    Secondly, since pixel samples have a fixed frame rate, there is no inherent pixel averaging of stable video regions across multiple frames, causing high-contrast and high-speed scenes to suffer greatly from noise in the dark regions of the video. Even the advent of extremely high frame rate cameras did not bring with it a new paradigm of video representations \cite{high_fps_hevc}. Rather, these cameras produce large, spatio-temporally redundant framed video files using standard codecs. While some video container formats allow for variable frame rate video, the pixel samples remain temporally organized across frames \cite{vvc}, and software support for such videos is limited.
    
    Finally, most computer vision algorithms depend on a framed interpretation of the world, digesting video as a sequence of frames with convolution \cite{cv_survey}. Despite great progress in the efficacy of such systems for tasks like object recognition, object tracking, and scene reconstruction, it is disingenuous to suggest that their structures emulate the human eye's continuous and dynamic sensing \cite{survey}.

\subsection{Event Video}\label{sec:event_video}
    More recently, we have seen the adoption of asynchronous event-based cameras that aim to address the above weaknesses of frame-based video capture and representation. Rather than output frames, these sensors output time-based events using an Address Event Representation (AER), wherein an event expresses a discrete data point for a particular pixel location and point in time. Here, we examine several event-based cameras and their corresponding AER data formats.
    
    \subsubsection{DVS}
        Dynamic Vision Sensors (DVS) prioritize localized capture rate over global accuracy. DVS cameras (\cref{fig:dvs}) continuously measure the log intensity, $\widetilde{L} = \ln(I)$, incident upon a pixel $\langle x,y\rangle$, until the brightness changes  such that $\widetilde{L}(x, y, t) - \widetilde{L}(x, y, t- \Delta t) = p\theta$, where $p \in \{-1,+1\}$ is the polarity of the brightness change, $\theta > 0$ is the threshold, and $\Delta t$ is the time elapsed since the pixel last recorded such a brightness change \cite{survey}. When the brightness changes across the threshold $p\theta$, the pixel outputs an asynchronous \textit{event}, represented by the tuple $\langle x, y, p, t\rangle$. The parameter $\theta$ is determined by a variety of camera gain settings, together with the effects of scene illumination. It is thus not easily determinable by the camera user, and may change drastically during the course of a video recording.
        

        The DVS approach has significant advantages. First, change events are quickly fired by pixels without incurring the frame latency of a synchronous sensor.  Second, a high dynamic range is achieved due to logarithmic compression of the intensity.  Third, when the intensity at a pixel is relatively stable, few events are fired, thereby conserving bandwidth. The key weaknesses of DVS are poor pixel noise characteristics and the difficulty of reconstructing actual intensity values.  The pixel values suffer from noise because log conversion is performed by a subthreshold MOS transistor, and the sensed value is an instantaneous voltage sample as opposed to an accumulation over an integrating interval.  Due to the differencing nature of sensing, the high-quality reconstruction of actual intensity values becomes difficult as noise accumulation leads to drift. Because of this shortcoming, DVS sensors are typically used when capturing the edges of moving objects is of primary importance.  This primarily includes robotics applications where fast response times are necessary and human visual quality is not important, such as object detection and tracking, gesture recognition, and SLAM (simultaneous localization and mapping). On the other hand, numerous learning-based \cite{Rebecq19pami,Rebecq19cvpr,Scheerlinck20wacv} and direct \cite{Scheerlinck,reinbacher,async_kalman_filter} methods for framed reconstructions of DVS data are proposed in the literature, but these are limited to reconstructing 100-1000 frames per second even on state-of-the-art hardware and low-resolution ($240\times 180$ pixels) sensors \cite{Rebecq19pami}.

    \subsubsection{DAVIS and ATIS}
    
        Dynamic and Active Pixel Vision Sensors (DAVIS) aim to improve the usability of DVS technology by co-locating active (APS) pixels alongside the DVS pixels on the sensor chip \cite{survey}. These active pixels capture discrete frames at a fixed rate, just as a traditional camera. APS data can improve the quality of intensity reconstructions by preventing drift in DVS accuracy \cite{Scheerlinck,Pan_EDI,high-framerate_recon,inherent_compression}. The resulting intensity reconstructions can exhibit very high dynamic range, and are useful in vision applications where broader scene understanding is important.
        
        Asynchronous Time-based Image Sensors (ATIS) take a different approach than DAVIS for capturing intensities alongside DVS events. Integrator subpixels on these sensors express absolute intensities by outputting two AER-style events, where a short time between the events corresponds to a bright pixel while a long time between events corresponds to a dark pixel \cite{survey}. Although these intensity measurements are accurate, they trigger only when the DVS subpixel records an intensity change event \cite{atis}. Thus, ATIS does not provide absolute intensities for every pixel at a given point in time, but only at high-contrast moving edges.
        
        ATIS and DAVIS aid vision tasks, such as object detection, by augmenting contrast-based events with absolute intensity measurement. However, they produce two distinct streams of data, and require the development of a sophisticated supporting software infrastructure to fuse the two streams to fully take advantage of their capabilities. For DAVIS in particular, the intensity frames help one to focus the camera's lens, use classical vision algorithms alongside event-based algorithms for comparison or augmentation, and achieve better performance in intensity reconstruction through the fusion of event and framed data. DAVIS has emerged as a prominent event-based sensing technology for research, while ATIS has not seen widespread commercial adoption. In this paper, we focus our attention on the DAVIS sensor.

        \begin{table*}[t]
  \caption{Comparison of event video representations. The \eventformat{} representation in transcode mode (ii) achieves the most flexibility for event-based applications, while it can also be trivially transformed to a frame-based intensity representation for use in classical applications.}
  \label{tab:representations}
  \begin{tabular}{p{2.8cm}|p{2.6cm}p{2cm}p{2cm}p{2cm}p{2cm}p{1.5cm}}
    \toprule
    Name &Representation & Continuous? & Full-sensor intensities? &Transformable to DVS? &Compatible with CNNs? &Compatible with SNNs?\\
    \midrule
    DVS & $\langle x,y,p,t \rangle$  & \cmark &   & — &   & \cmark\\
    DAVIS & $\langle x,y,p,t \rangle,$ $[w, h, I]$  & DVS only & APS only & — & APS only & DVS only \\
    DAVIS image fusion & $[w, h, I]$  &   & \cmark &   & \cmark &   \\
    DVS event frames & $[w, h, \Delta L]$  &   &   &   & \cmark &  \\
    DVS time surfaces & $[w, h, t]$ &   &   &   & \cmark &  \\
    DVS voxel grid & $[w,h,n,$ $\sum p\text{ or }\frac{\partial L}{\partial t}]$ &   &   &   & \cmark &   \\
    ATIS & $\langle x,y,p,t \rangle,$ $\langle x,y,I,t \rangle$ & \cmark &   & — &   & \cmark \\ \hdashline
    \eventformat{} event frames & $[w, h, \frac{2^D}{\Delta t}]$ &   & \cmark &   & \cmark &   \\
    DAVIS (i) $\rightarrow$ \eventformat& $\langle x, y, D, \Delta t\rangle$ & \cmark & \cmark &  &   & \cmark\\
    DAVIS (ii) $\rightarrow$ \eventformat& $\langle x, y, D, \Delta t\rangle$ & \cmark & \cmark & \cmark   &   & \cmark\\
    DVS (iii) $\rightarrow$ \eventformat& $\langle x, y, D, \Delta t\rangle$ & \cmark &   & \cmark &   & \cmark\\
    \bottomrule
  \end{tabular}
\end{table*}
        
    \subsubsection{Event Video Representations}\label{sec:event_representations}
          While some applications process DVS events directly with spiking neural networks (SNNs) \cite{Duwek_2021_CVPR,Barbier_2021_CVPR,spiking1}, many resort to grid-based representations of the events, which are more amenable to classical vision pipelines and do not require specialized neuromorphic network hardware \cite{Gehrig_2019_ICCV,eventframe,eventcar,EV-FlowNet,eventflow,TORE_volumes}. We outline several of these grid-based representations in \cref{tab:representations}, highlighting the tradeoffs researchers make between temporal precision and ease of processing.
          
          Although researchers have explored rate control schemes for DVS streams, the proposed systems perform rate adaptation only at the application level \cite{glover}, or at the camera source by adjusting biases for $\theta$ \cite{dvs_feedback_control,dvs_rate_patent}, rather than exploring a rate-distortion control scheme for event representation and transmission. In a DAVIS camera, the stream of APS frames is a separate representation from the DVS event stream. Researchers may fuse the active and dynamic information as discussed above, but often still fall back to framed representations to incorporate the disparate streams. This approach temporally quantizes the events, discarding much of the high-rate advantage of DVS capture. While Banerjee et al. uniquely propose a system for application-driven DAVIS rate control on a networked client \cite{quadtree_compression,banerjee2021joint}, their scheme does not unify the event and the framed information under a single representation, meaning that the streams still require separate application-level logic.
         
         Additionally, the application ecosystem for event vision is fragmented between numerous raw data representations. Data sets are variously distributed in a human-readable text format \cite{davis_dataset,v2e}, a \texttt{.bag} format for the Robot Operating System \cite{davis_dataset,Rebecq19pami,lee2022vivid,v2e}, an \texttt{.hdf5} format for learning-based Python tasks \cite{ddd20,v2e}, an \texttt{.aedat2} format designed for operability with the jAER software \cite{ddd20,v2e}, or a \texttt{.mat} format for use in MATLAB applications \cite{Pan_EDI,async_kalman_filter}. Event camera manufacturers iniVation and Prophesee each have their own software suites providing applications, but each operates only on their proprietary camera data formats. While some tools exist to transform raw camera data to one of the above intermediate representations, none of them fundamentally changes the underlying data types. We argue that, if one must already transform the raw data to prepare it for an application, we should instead transcode it to a simple, fast, compressible, non-proprietary data representation which is both application agnostic and camera agnostic. Then, applications may ingest this single representation, and support for future event sensing modalities can be implemented in the data transcoder rather than in individual applications. 
    

\section{Proposed Representation}

\subsection{Motivation}
 While we recognize the utility of a DVS-style representation for contrast-based events, such a format is not amenable to representing absolute intensities.
With a DVS-style approach, inducing loss is feasible only in terms of mitigating sensor noise and allowing an application to ignore events it deems unimportant. That is, removing a DVS event in a given pixel's stream will necessarily affect the interpretation of later events in that pixel's stream, since each event conveys intensity information \textit{relative} to a previous event. For the same reason, even the DVS-style ATIS representation, which can encode intensity information, is ill-suited as a general, adaptable, and compressible representation. Specifically, these representations do not support receiver-driven lossy compression of their events.

 In response, we propose the \textbf{A}ddress, \textbf{D}ecimation, $\Delta t$ \textbf{E}vent \textbf{R}epresentation (\textbf{\eventformat{}}, pronounced “adder'') format as the “narrow waist'' representation for asynchronous video data. This representation draws from a proposed Asynchronous Integration (ASINT) camera model \cite{montek,Smith2017ASM,freeman_emu}. As illustrated in \cref{fig:asint}, a pixel $\langle x,y,c\rangle$ continuously integrates light, firing an \eventformat{} event $\langle x,y,c,D,\Delta t\rangle$ when it accumulates $2^D$ intensity units (e.g., photons), where $D$ is a \textit{decimation threshold} and $\Delta t$ is the time elapsed since the pixel last fired an event. We measure $t$ in clock ``ticks,'' where the granularity of the clock tick length is user adjustable. Unlike ATIS events, a \textit{single} \eventformat{} event directly specifies an intensity, $I$, by
        $I = \frac{2^D}{\Delta t}$.
    The key insight of this model is \textbf{the dynamic, pixelwise control of $D$} during transcode. Raising $D$ for a pixel will decrease its event rate, while lowering $D$ will increase its event rate. With this multifaceted $D$ control, we can ensure that pixel sensitivities are well-tuned to scene dynamics, avoiding the rate and compression issues of the ATIS model while maintaining the strengths of intensity events.
    
    Although related work focused on emulating the proposed ASINT sensor and developing early compression schemes and applications \cite{freeman_emu,FreemanLossyEvent,Freeman2021mmsys}, here we distinguish the function of the sensor from its representation. That is, we look to buffered processing techniques to encode optimal \eventformat{} representations for an arbitrary set of inputs, and we posture \eventformat{} as a transcoded representation for \textit{non}-ASINT sources. With this representation, we have  a  unified  format  for which to build compression schemes and applications targeting a wide variety of camera technologies.  With source-specific transcoder modules and an abstract \eventformat{} library, we can cast video from arbitrary camera sources (both framed and non-framed) into the single, unified \eventformat{} data representation.
    

\subsection{\eventformat{} Transcoder Fundamentals}\label{sec:transcoder_fundamentals}
    
    To explain how we transform data into \eventformat, we must first describe our novel pixel model and parameters common to all transcode types.

    \subsubsection{Event Pixel List Structure}\label{sec:pixel_list_structure}
    We will first examine the integration of a single \eventformat{} pixel model. We represent an \eventformat{} pixel state with a linked list. Each node in the list consists of a decimation factor, $D$, an intensity integration measurement, $I$, and a time interval measurement, $\Delta t$. The connections between nodes carry an \eventformat{} event representing the intensity that the parent node integrated \textit{before} the child node was created.  When initializing a new list, we set $D$ to be the maximal value possible to represent the first intensity we intend to integrate. To maintain precision, the values $I$ and $\Delta t$ in the list are floating point numbers. Since \eventformat{} is source-agnostic, we interpret the incoming values as generic ``intensity units'', rather than tying them to a real-world measure of intensity, such as photons, and we can scale the input values of individual data sources according to our needs.
    
\begin{figure}[h]
        \centering
        \includegraphics[width=1.0\linewidth]{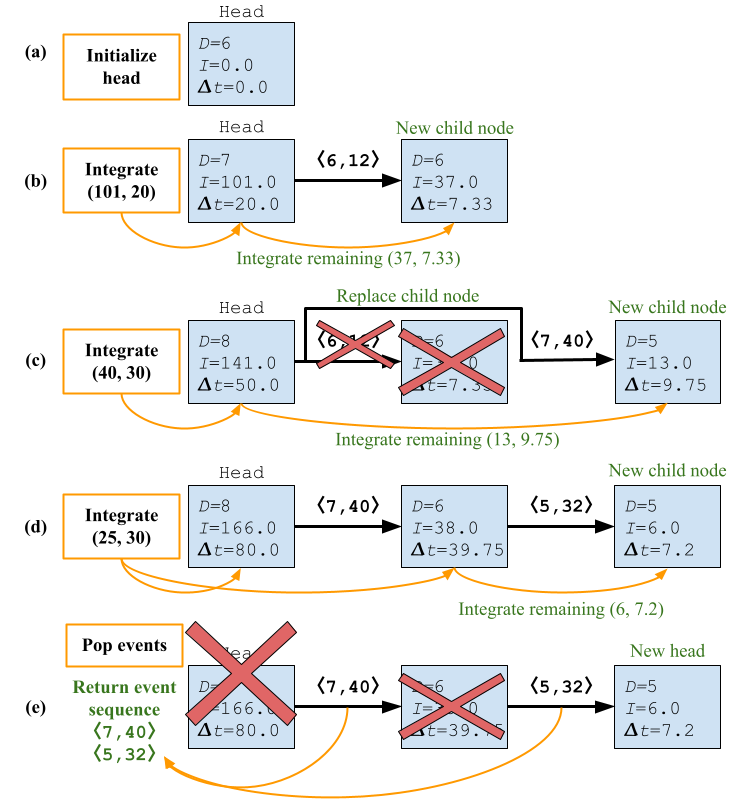}
        \caption{Structure of event pixel lists under continuous integration.}
        \label{fig:pixel_list}
    \end{figure}

    Let us walk through a visual example of this list structure in \cref{fig:pixel_list}. Suppose our first intensity to integrate is 101. We initialize our head node with $D=\lfloor\log_2(101)\rfloor = 6$ in \cref{fig:pixel_list}a. 
    
    Now, suppose we integrate the intensity 101, spanning 20 time units (ticks). The head only accumulates $64/101$ of the intensity units before saturating its $2^D=64$ integration. Thus, the time spanned for this partial integration is $(64/101)\cdot 20 = 12.67$ ticks. We can represent an \eventformat{} event (without its spatial coordinate), illustrated in angle brackets, as the node connection in \cref{fig:pixel_list}b. These events consist of integer numbers, so we use $\lfloor\Delta t \rfloor = 12$. At this stage, we create a child node to represent the remaining integration. The child takes on the head node's $D$ (i.e., $D=6$).  We integrate the remaining $101-64=37$ intensity units for the child node, spanning $(37/101)\cdot 20 = 7.33$ ticks. Finally, we increment the head node's $D$ value and integrate the remaining $I=37, \Delta t = 7.33$ intensity.
    
    Let us now integrate 40 intensity units over 30 ticks, as in \cref{fig:pixel_list}c. The head node saturates its $2^D=128$ integration, so we \textit{replace} the child node with a new node, connected by the new event $\langle D=7, \Delta t = \lfloor 20 + (27/40)\cdot 30\rfloor = 40 \rangle$. In this way, we can minimize the number of \eventformat{} events required to represent an intensity sequence, maximizing the $D$ and $\Delta t$ values of the events. As before, we integrate the remaining intensity after spawning the child to both the head and the child nodes.

    Finally, let us integrate 25 intensity units over 30 ticks, as in \cref{fig:pixel_list}d. We first integrate the head event. It does \textit{not} reach its $2^D=256$ integration threshold, so we do not replace its event nor replace the child node. Then we integrate the child node with the same intensity, reaching its $2^D=32$ threshold with  $19$ intensity units and an additional $(19/25)\cdot 30 = 22.8$ ticks. Thus, we increment $D$, create an event for the child, $\langle D=5, \Delta t = \lfloor 9.75 + (19/25)\cdot 30\rfloor = 32 \rangle$, and spawn another node in the list to integrate the remaining $6$ intensity units across $(6/25)*30=7.2$ ticks.
    
    The node connections are a queue which a transcoder outputs when there is a significant change in the intensities being integrated. When such a criterion is met, we can dequeue the events from the list, and the tail node becomes the new head, as in \cref{fig:pixel_list}e. This linked list structure minimizes the number of \eventformat{} events required to represent a sequence of intensity integrations and eliminates the ``speckle'' artifacts present in the work of Freeman et al. \cite{freeman_emu,FreemanLossyEvent,Freeman2021mmsys}.
    
    \subsubsection{User-Tunable Parameters}\label{sec:user_parameters}
    A user may set a number of trans\-code parameters according to their needs. 
    
    $\Delta t_{s}$: Number of ticks per second. This parameter defines the temporal resolution of the \eventformat{} stream. 
    
    $\Delta t_{ref}$: Number of ticks for a standard length integration (e.g., an input frame exposure time, when the source is a framed video). This parameter, along with $\Delta t_{s}$, determines the accuracy and data rate.
    
    $M$: \eventformat{} contrast threshold. When a pixel's incoming normalized intensity exceeds its baseline intensity by this threshold (either positively or negatively), then the pixel's event queue will be output, its integration state reset, and its baseline set to the new intensity. Unlike the DVS model, which detects changes in log intensity, we look at the change in absolute intensity. In this paper, $M$ is uniform for every pixel, although future applications can dynamically adjust $M$ per-pixel with no added complexity. Below, we evaluate the effect of $M$ on transcoded quality and event rate.
    
    \dtm: The maximum $\Delta t$ that any event can span. Suppose that there is a static scene; this parameter is directly correlated with the event rate. That is, halving \dtm{} would result in a doubling of the event rate. In practice, however, the user will simply want to ensure that \dtm{} is sufficiently large enough so that the desired amount of intensity averaging will occur in temporally stable regions of the video, but small enough so that the given application will receive pixel updates at a fast enough rate.
    
    \subsubsection{Binary Representation}\label{sec:binary_representation}
    A raw \eventformat{} file begins with a header containing metadata that describe the resolution of the video, the file specification version, the endianness, $\Delta t_{s}$, $\Delta t_{ref}$, and $\Delta t_{max}$. Immediately after this header, there is a sequence of raw \eventformat{} events in the following format.
    
    \textit{x}: unsigned 16-bit \textit{x} address.
    
    \textit{y}: unsigned 16-bit \textit{y} address.
    
    \textit{c}: optional unsigned 8-bit color channel address. If the video is monochrome, then the $c$ value is absent.
    
    $D$: unsigned 8-bit decimation value.
    
    $\Delta t$: time spanned since the last event of this pixel.
    
    We use a packed representation, so that each event is 9-10 bytes, depending on the presence of the color channel. However, we emphasize that this style of event representation is highly compressible, as demonstrated in \cite{FreemanLossyEvent,Freeman2021mmsys}. We note that the range of natural values for $D$ is $[0,127]$, since $2^{127}$ is the maximum representable unsigned integer in our language of choice, although $D$ values will typically span the range $[0,30]$. We additionally reserve $D=254$ as a special symbol that indicates that the event represents a 0-intensity, which cannot be communicated with $D\in[0,127]$ alone.
    
    
    \subsubsection{Implementation Details}\label{sec:implementation_details}
    
    We implemented our entire framework in the Rust language. Non-Rust dependencies include OpenCV \cite{opencv_library} for image ingest, processing, and visualization, and FFmpeg \cite{ffmpeg} for framed video coding. The framework is fast, highly parallel, and optimized to take advantage of CPU resources. For all speed evaluations in this paper, we used an 8-core AMD Ryzen 2700x CPU, with output files written to a RAM disk.
    
\section{Framed Video Sources}\label{sec:framed_video_sources}
We begin by describing how we apply our \eventformat{} pixel model to framed video sources.

\subsection{Transcoder Details}\label{sec:framed_transcoder_details}
We conceptualize a video frame as a matrix of intensities integrated over a fixed time period, assuming that the intensity for each pixel may change only at instantaneous moments between frames (\cref{fig:framed_diagram}). $\Delta t_{ref}$ defines the integration time of each frame, and we set $\Delta t_{s} = \Delta t_{ref}F$ ticks for the source video frame rate, $F$. Since each \eventformat{} pixel integrates exactly one intensity per input frame, we can easily parallelize this process by integrating groups of pixels on many CPU threads, then collecting the resulting \eventformat{} events into a single vector to write out after integrating each frame. Since each pixel is processed independently in the \eventformat{} model, temporally interleaving the events of different pixels is unnecessary. For example, Pixel A may fire events $A_1$, then $A_2$, while Pixel B may have an event $B_1$ that falls between the times of $A_1$ and $A_2$. In this case, it is valid to encounter $A_1$ and $A_2$ before $B_1$ in the \eventformat{} stream, but we will never encounter $A_2$ before $A_1$.

\subsubsection{Framed Reconstruction and Preserving Temporal Coherence}\label{sec:framed_reconstruction}
To easily view and evaluate the effects of our \eventformat{} transcode with traditional methods, we can perform a framed reconstruction of the data. For this, we simply maintain a counter of the running timestamp, $T$ for each pixel in the \eventformat{} stream. When reading an event $\langle x,y,D', \Delta t'\rangle$ for a given pixel, we normalize the intensity per frame to $I_{frame} =\frac{2^{D'} \Delta t_{ref}}{\Delta t'}$, then set this as the pixel value for all frames in $[T/\Delta t_{ref}, (T+\Delta t')/\Delta t_{ref}]$.

We must, however, ensure that intensity changes only occur along frame temporal boundaries. Otherwise, there would be temporal decoherence as certain pixels fire slightly earlier than others, unless the source video is recorded at an extremely high frame rate. To this end, we ignore any intensity remaining for a given pixel list state after it generates a new \eventformat{} event, thus ensuring that the temporal start of each event corresponds to the beginning of a frame in the source. We can simply infer the time between events when processing the events, by rounding $T$ up to the next multiple of $\Delta t_{ref}$. Our framed reconstruction program handles this case automatically, since both the source data type and $\Delta t_{ref}$ are encoded in the header of the \eventformat{} file.

\begin{figure}[h]
        \centering
        \includegraphics[width=0.83\linewidth]{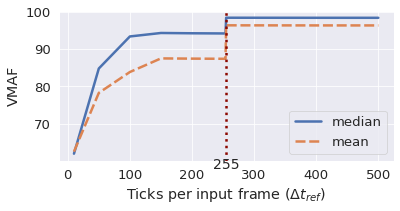}
        \caption{Effect of $\Delta t_{ref}$ on VMAF perceptual quality of \eventformat{} framed reconstructions. Source videos are 8-bit PCC.}
        \label{fig:tpf_vmaf}
    \end{figure}
    
\begin{figure*}[t]
    \centering
    \subfloat[ ]{\includegraphics[width=0.25\linewidth]{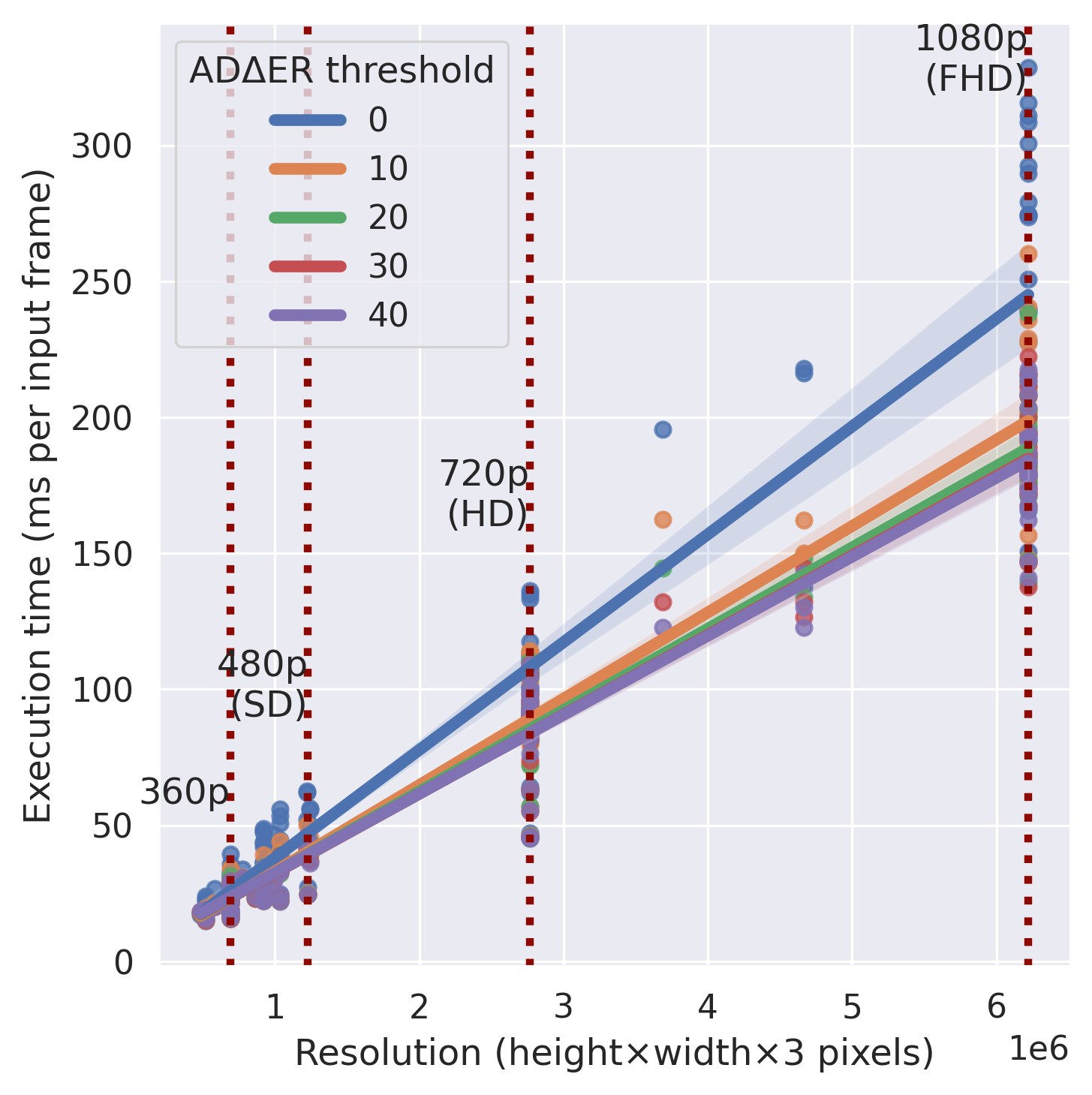}%
    \label{fig:framed_execution_times}}
    \hfil
    \subfloat[ ]{\includegraphics[width=0.25\linewidth]{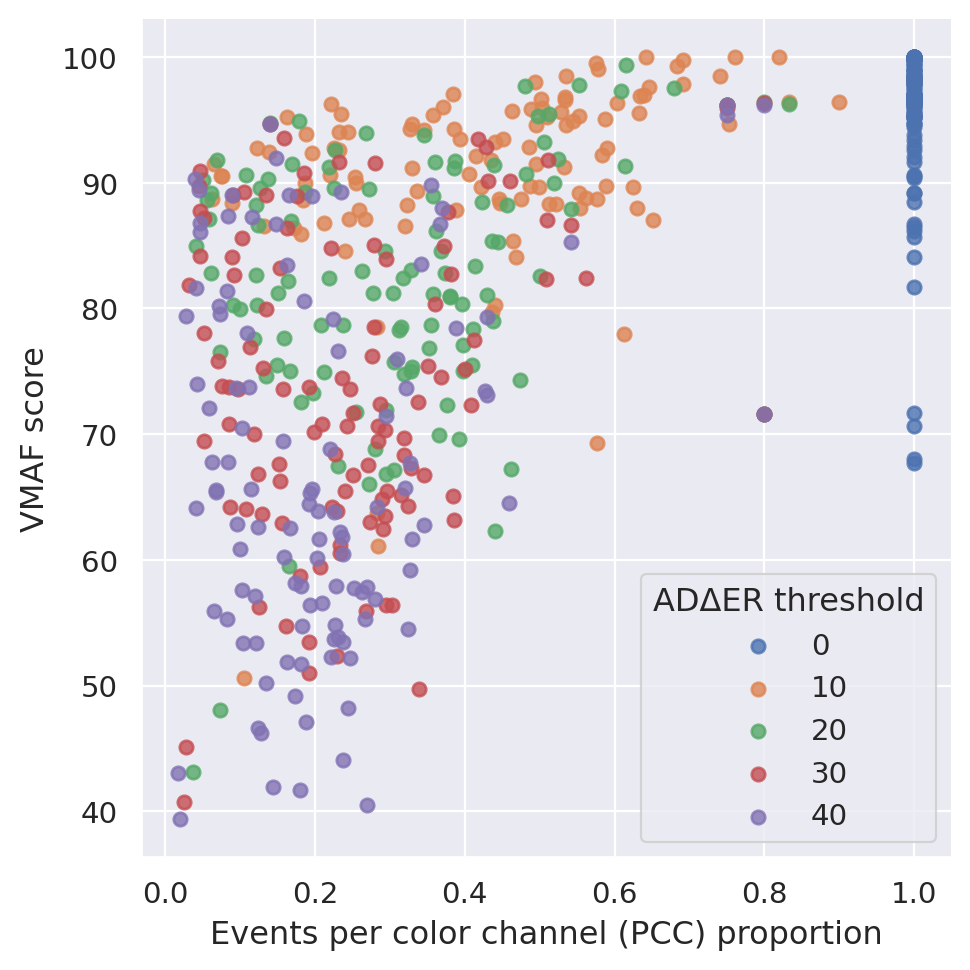}%
    \label{fig:framed_event_rate_vmaf}}
    \hfil
    \subfloat[]{\includegraphics[width=0.25\linewidth]{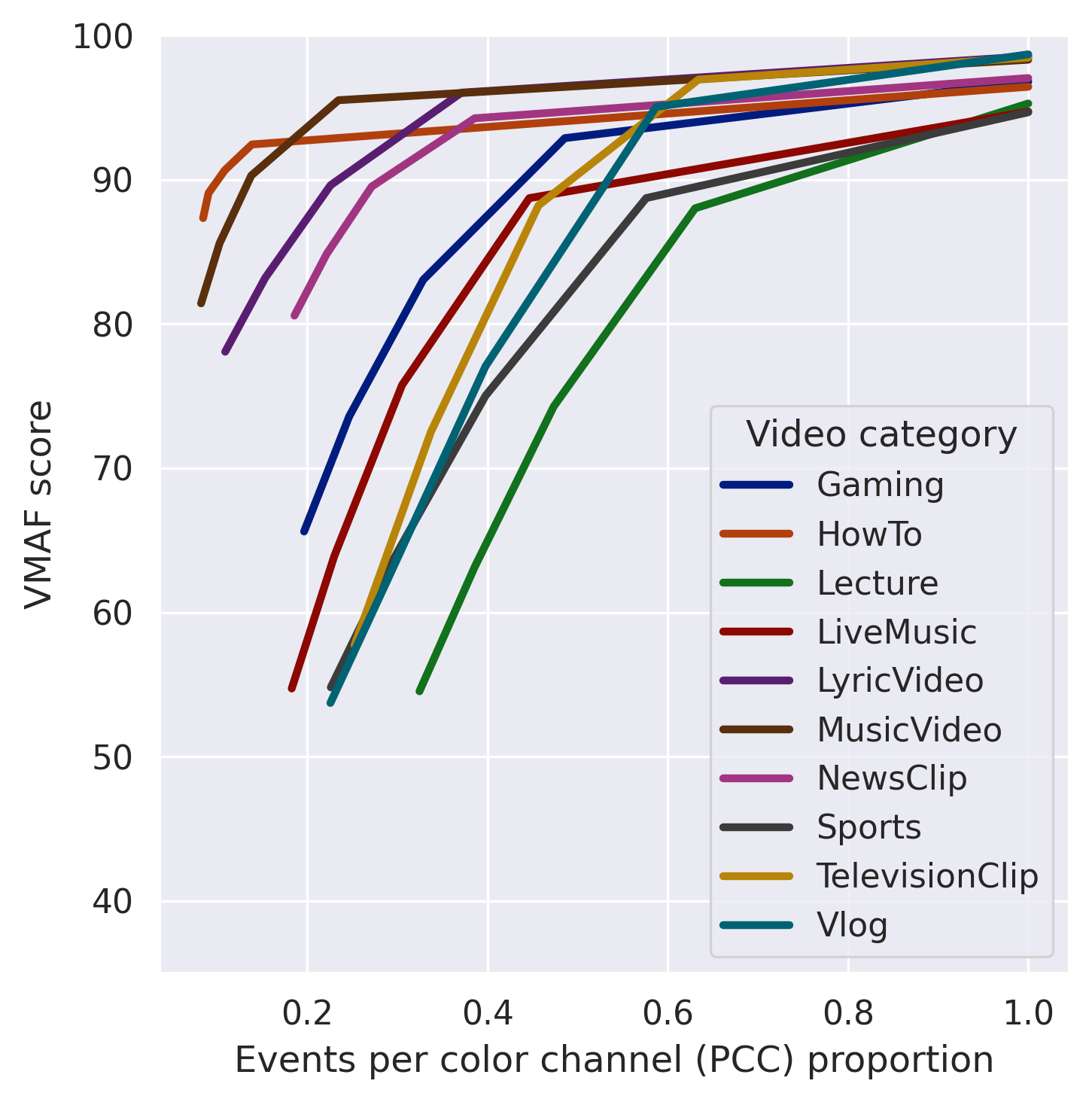}%
    \label{fig:framed_event_rate_vmaf_particular}}
    
    \caption{Quantitative results of \eventformat{} on framed sources. (a) Effect of resolution on execution time, at various $M$. Performance scales linearly with resolution. (b) Scatter plot of the rate-distortion for all videos at various $M$. (c) Particular rate-distortion curves for one video in each category of the data set. Increasing $M$ yields lower-quality, but lower-rate \eventformat{} videos.}
    \label{fig:framed_eval}
\end{figure*}

\subsection{Optimizing $\Delta t_{ref}$}
Since we define $\Delta t_{s}$ relatively to $\Delta t_{ref}$, it is crucial to understand the effect of $\Delta t_{ref}$ on the transcoded representation's quality. We transcoded a diverse set of 10 framed videos to \eventformat{} with $M=0$ (for the most accurate representation) at various choices of $\Delta t_{ref}$. We then performed framed reconstruction of the \eventformat{} streams to use the VMAF \cite{vmaf} metric to calculate the perceptual quality of the reconstructions compared to the source videos, as illustrated in \cref{fig:tpf_vmaf}. We see that at least 255 ticks per input frame is a necessary parameter for strong reconstruction quality, due to the \textit{bit depth} of the source videos (namely, 8 bits per channel). The maximum intensity of the source frame is $\frac{255}{255} =1$ intensity units per tick in $\Delta t_{ref} = 255$, which we can represent with $\langle D=7, \Delta t =128\rangle$. The minimum intensity is 0, which we may represent with $\langle D=254, \Delta t =255\rangle$. However, if $\Delta t_{ref} = 254$, we see that the maximum intensity $\frac{255}{254} =1.004$ is not representable with an integer $\Delta t$ value. Thus, we must set $\Delta t_{ref}$ large enough that we may represent any source intensity without losing accuracy, such that $\frac{I_{max}}{\Delta t_{ref}} \leq 1.0$.

\subsection{Evaluation}\label{sec:framed_evaluation}
\begin{figure}[h]
        \centering
        \includegraphics[width=0.9\linewidth]{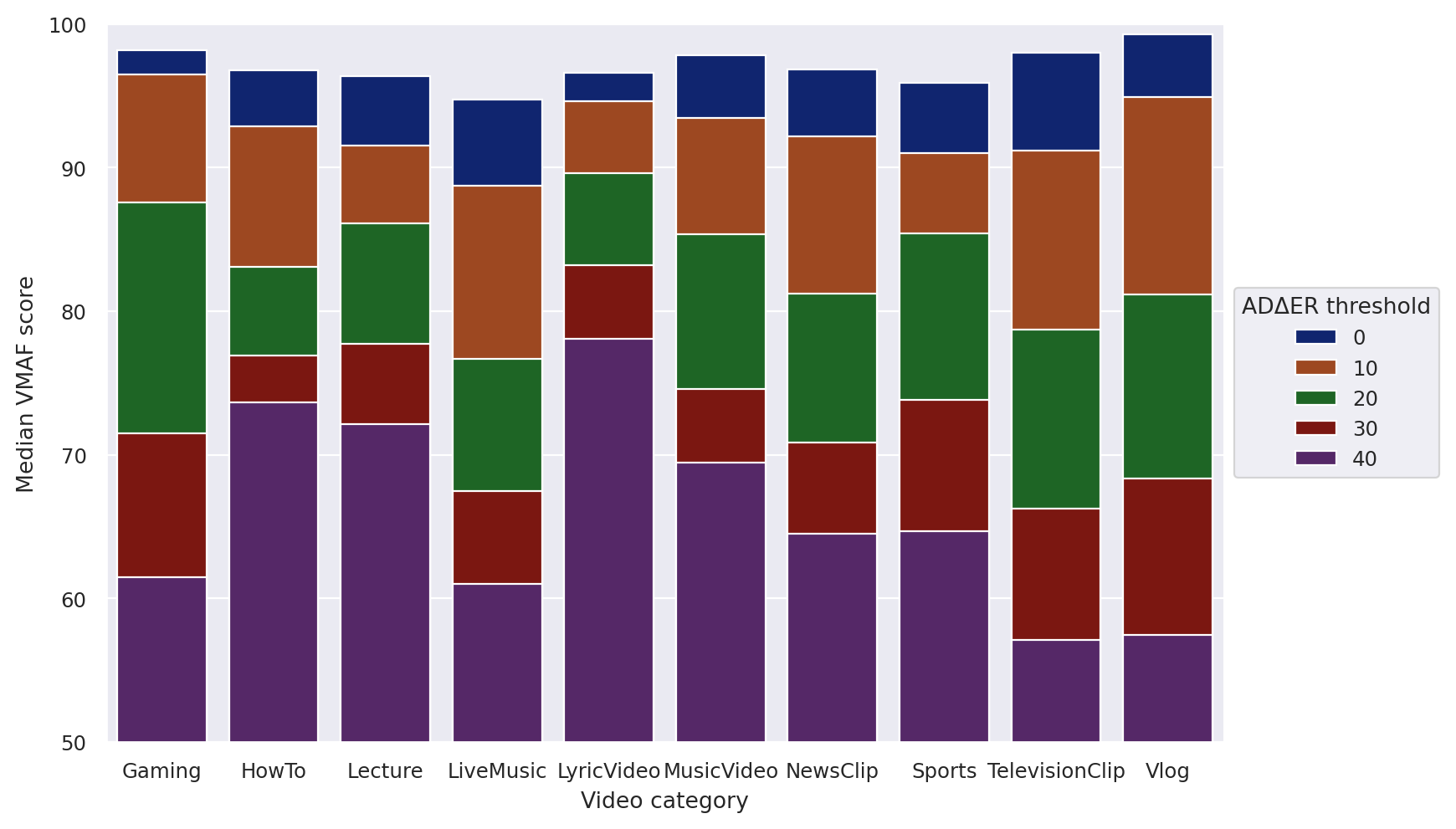}
        \caption{Median VMAF score for each framed video category after transcoding to \eventformat{} at various $M$. Quality decreases as $M$ increases for every category. Less dynamic video categories such as \texttt{LyricVideo} have little quality degradation as $M$ increases, while more dynamic categories such as \texttt{TelevisionClip} and \texttt{Vlog} degrade greatly.}
        \label{fig:median_vmaf}
    \end{figure}

To evaluate our \eventformat{} representation on framed video sources, we employed 112 videos from a subset of the YT-UGC video compression data set \cite{yt-ugc-dataset}. These 20-second videos span 10 categories (listed in \cref{fig:median_vmaf}) and have resolutions 360p, 480p, 720p, and 1080p. We used the variants of the videos provided with H.264 compression at the CRF-10 quality level, and each video has 24-bit color.

We transcoded each video to \eventformat{} with parameters $\Delta t_{ref} = 255$ and $\Delta t_{max} = \Delta t_{ref} \cdot 120 = 30600$, such that the longest $\Delta t$ representable by any generated \eventformat{} event can span 120 input frames. As described in \cref{sec:framed_transcoder_details}, we set $\Delta t_{s} = \Delta t_{ref}F$ for each video. For example, a 24 FPS video has $\Delta t_{s} = 6120$. For each video, we varied the \eventformat{} threshold parameter $M \in \{0, 10, 20, 30, 40\}$. We recorded the execution time on our test machine (\cref{sec:implementation_details}) for each transcode, as illustrated in \cref{fig:framed_execution_times}. We see that larger values of $M$ yield slightly faster transcode operations, and that the time required to process an input frame scales linearly with video resolution.
    
Finally, we reconstructed framed videos from our \eventformat{} representations as described in \cref{sec:framed_reconstruction}, so that we can examine the quality properties of our representation. We gathered the VMAF score of each \eventformat{} video in the $M$ range, compared to its reference video. \cref{fig:median_vmaf} shows that the median VMAF score for each video category is extremely high, in the 95-99 range, and that perceptual quality decreases as $M$ increases.

If we examine the number of \eventformat{} events per pixel color channel at various levels of $M$ as a proportion compared to $M=0$  (\cref{fig:framed_event_rate_vmaf,fig:framed_event_rate_vmaf_particular}), we see a clear rate-distortion curve dependent on the choice of $M$. By merely setting $M=10$, we see a median reduction in \eventformat{} rate by $54\%$, and a median reduction of VMAF score by only 4.5 across our data set. 
In our tests, we observe a maximum \eventformat{} precision of 44 dB or 14 bits, demonstrating a significant increase over the 8-bit precision of the source videos. The precision tends to increase as pixels can integrate (and thus average) longer sequences of intensities with higher $M$, thus widening the range of represented values.

As a qualitative example, \cref{fig:framed_matrix} shows instantaneous frame samples from the \eventformat{} transcodes with our range of $M$ values\footnote{We provide the framed reconstructions of these videos as supplementary material \href{https://drive.google.com/drive/folders/1VXAMPrNnm0qRDqyz0pqDV_dJTNh-h_hm?usp=sharing}{here}.}. With higher $M$, we observe a greater ``smearing'' effect in low-contrast regions, less overall detail, and multicolored ghosting artifacts from scene transitions or fast camera motions. Surprisingly, the \texttt{LiveMusic} and \texttt{Sports} examples shown are both outliers for the VMAF score with $M=0$ in \cref{fig:framed_event_rate_vmaf}, despite appearing high quality to a casual observer. This is due to the imprecision of some events with $\Delta t > \Delta t_{ref}$, near high-contrast transition points, since we round $\Delta t$ to an integer to form an \eventformat{} event. Furthermore, since we represent each color channel with an independent event pixel list, a small error in one color channel of a pixel will have a compound effect on the VMAF perceptual quality of that pixel. These two videos exhibit jittery camera motion, suggesting that rapid intensity changes may make such quantization errors more frequent.

\section{Event Video Sources}\label{sec:event_video_adder}
While in \cref{sec:framed_video_sources} we demonstrated that \eventformat{} can effectively represent framed video sources asynchronously, we here discuss \eventformat{}'s utility in representing video sources which are already asynchronous. Specifically, we explore the representation of the DAVIS 346 camera's DVS event and APS frame data in the \texttt{.aedat4} file format.

\subsection{Event-Based Double Integral}
As we discussed in \cref{sec:event_video}, many applications that utilize mixed-modality sources (frames and events) either reconstruct an image sequence at a fixed frame rate or process the frame data separately from the events. Furthermore, while APS frames provide absolute intensity measurements across the whole sensor, they can often be blurry in difficult exposure scenarios: low scene illumination, fast motion, or narrow aperture. Meanwhile, the higher sensitivity of DVS pixels and their fast responsiveness allow them to capture events with subframe precision. Thus, DAVIS-based reconstruction methods must address the crucial problem of \textit{deblurring} the APS frame data. Most of these efforts, however, utilize slow machine learning techniques which severely limit the practical reconstructed frame rate, obviating a primary benefit to using event cameras \cite{wang2020event,song_ecir}. One non-learned method for framed reconstruction, however, is the Event-based Double Integral (EDI) \cite{Pan_EDI}. The authors introduced the EDI model to find a sharp ``latent'' image, $L$, for a blurry APS image, by integrating the DVS events occurring during the APS image's exposure time. Crucially, this paper involves an optimization technique for determining the $\theta$ value of the DVS sensor at a given point in time; that is, the method deblurs an APS image with various choices of DVS sensitivity, $\theta$, and the sharpest latent image produced corresponds to the optimal choice of $\theta$. Both the APS deblurring and $\theta$ optimization techniques are vital steps in our \eventformat{} transcoder pipeline for event video sources.

\subsubsection{Our Implementation}

Unfortunately, as with much of the literature on event-based systems, the EDI model was not implemented with practical, real-time performance in mind. Rather, the authors released their implementation as an obfuscated MATLAB program, which can only reconstruct a few frames of video per second. Furthermore, their implementation uses a custom MATLAB format for the DAVIS data, underscoring our argument about the prevalence of domain-specific event data representations in \cref{sec:event_representations}. Therefore, we implemented the EDI model from the ground up in Rust, designing our program to decode data in the packet format produced directly by the DAVIS 346 camera. Our implementation can reconstruct a frame sequence for this resolution at 1000+ FPS, two orders of magnitude faster than the baseline MATLAB implementation. To allow for realistic and repeatable performance evaluation, we simulate the packet latency of pre-recorded videos to reflect their real-world timing. The user can also generate one deblurred image for each APS frame, rather than a high-rate frame sequence. Our implementation supports a connected DAVIS 346 camera as input, unlocking the potential for practical, real-time DAVIS applications.

\subsection{Transcoder Details}
\begin{figure}[h]
        \centering
        \includegraphics[width=0.81\linewidth]{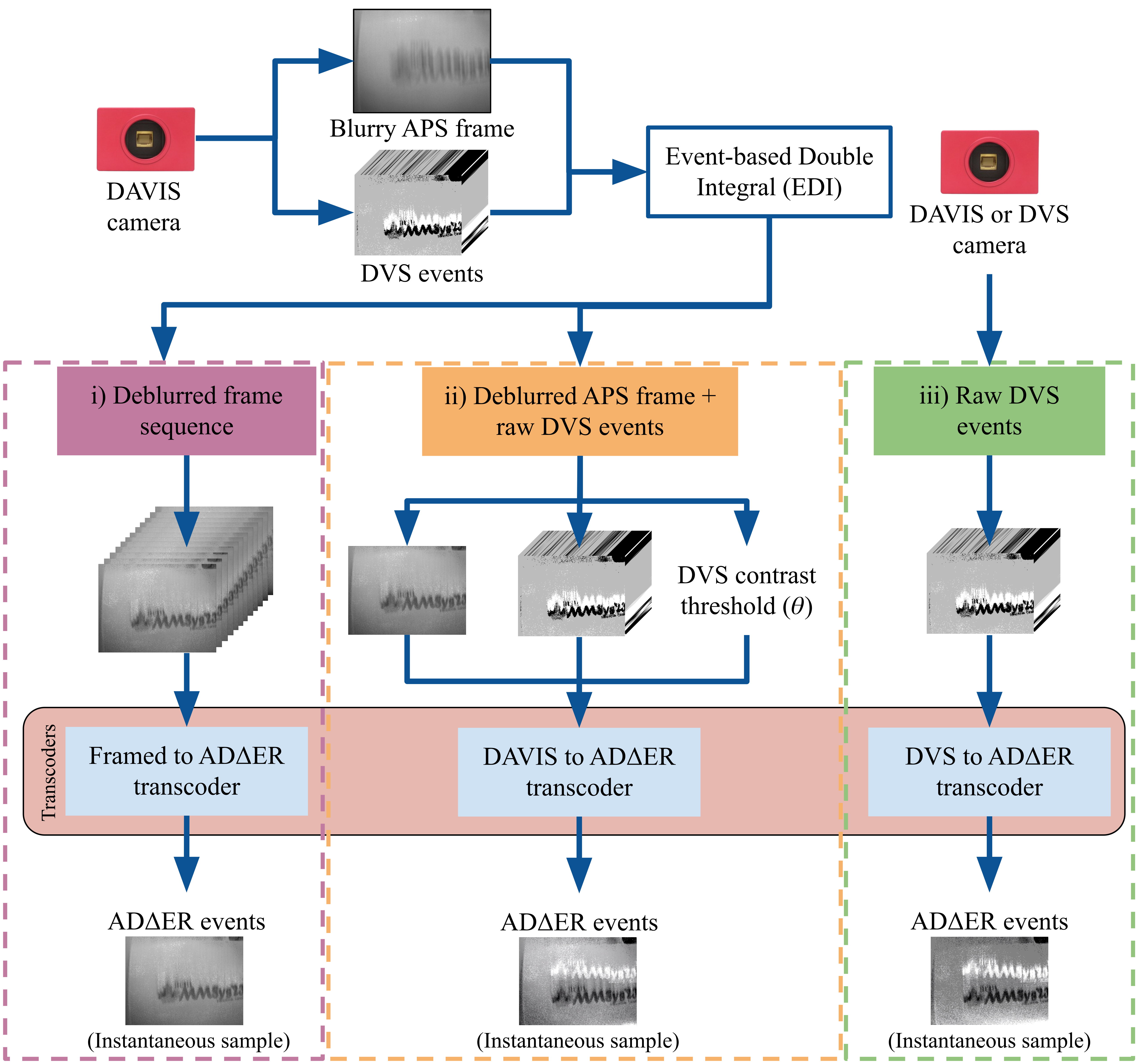}
        \caption{Pipeline for transcoding event camera data to \eventformat{}. Our transcoder supports three modes. Modes (i) and (ii) incorporate deblurred APS frames from a DAVIS sensor, while mode (iii) uses DVS events alone.}
        \label{fig:davis_pipeline}
    \end{figure}

We identify three ways in which we can transcode DAVIS data to \eventformat{}. Each mode utilizes the same per-pixel integration scheme as described in \cref{sec:transcoder_fundamentals}, but the actual intensities that we \textit{integrate} depend on the mode we choose. In each case, we receive input data from our EDI implementation running alongside our transcoder. Unlike our framed source transcoder, where we determine $\Delta t_{s}$ from our choice of $\Delta t_{ref}$ and the video frame rate, here we set $\Delta t_{s} = 1\times 10^6$ to match the temporal resolution of the DAVIS 346. Then, $\Delta t_{ref}$ defines the desired length of a deblurred frame in modes (i) and (ii) below, and the intensity scale in mode (iii) below. We illustrate our pipeline in \cref{fig:davis_pipeline}.

\subsubsection{Mode (i): Deblurred frame sequence $\rightarrow$ \eventformat}\label{sec:mode_1}
If we reconstruct a high-rate framed sequence with EDI, we can trivially use our framed-source transcoder technique as described in \cref{sec:framed_transcoder_details}. Qualitatively, this method produces the best-looking results, since the EDI model inherently unifies positive- and negative-polarity events in log space. That is, we can accumulate a pixel's multiple events occurring over a given frame interval to arrive at a final intensity value for the given frame. Any intensities outside the APS frame's intensity range $[0,255]$ are clamped only at this point.

\begin{figure*}[t]
    \centering
    \subfloat[]{\includegraphics[width=0.21\linewidth]{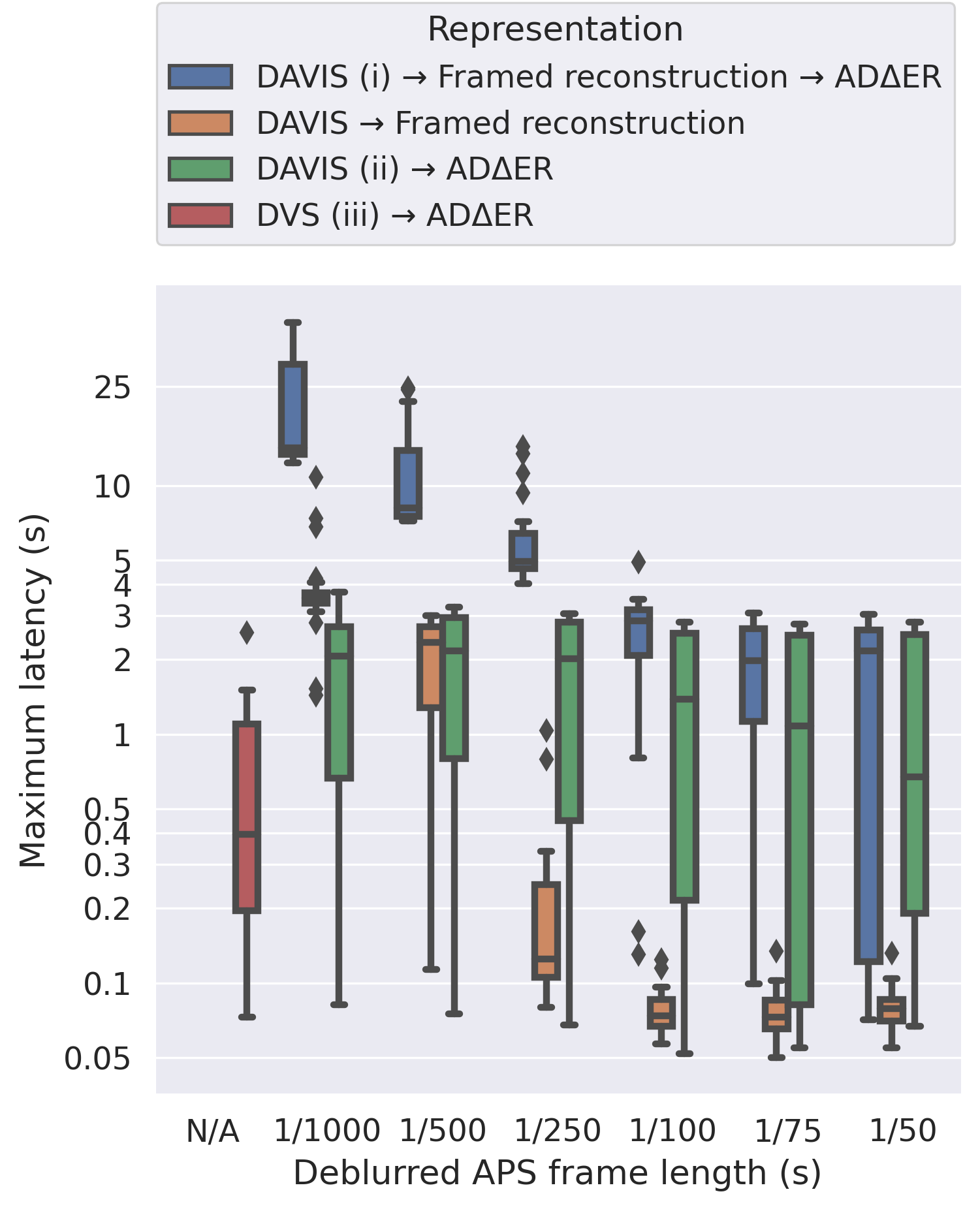}%
    \label{fig:davis_allmodes_latency}}
    \hfil
    \subfloat[]{\includegraphics[width=0.235\linewidth]{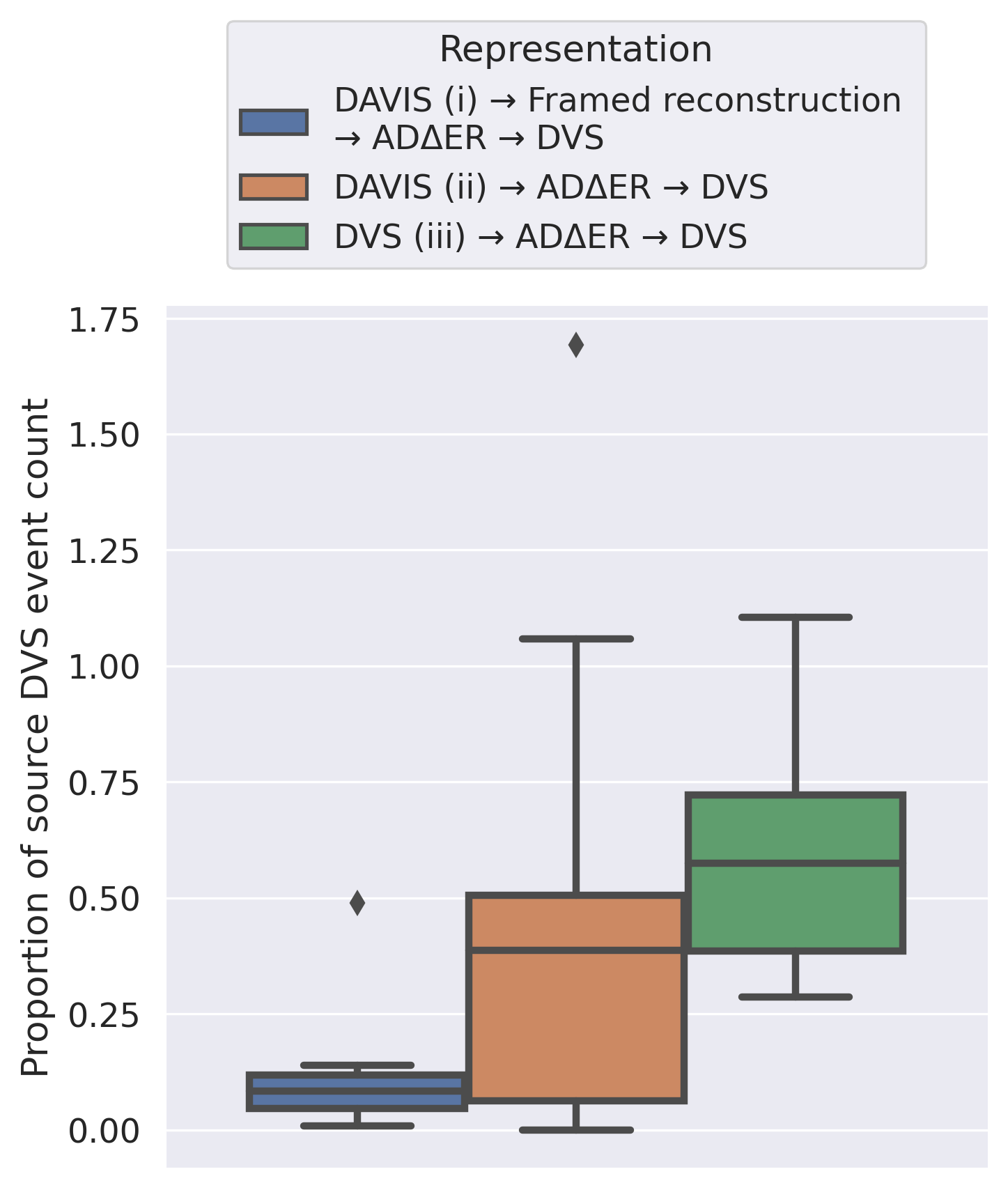}%
    \label{fig:back_to_dvs_event_rates}}
    \hfil
    \subfloat[]{
    \includegraphics[width=0.44\linewidth]{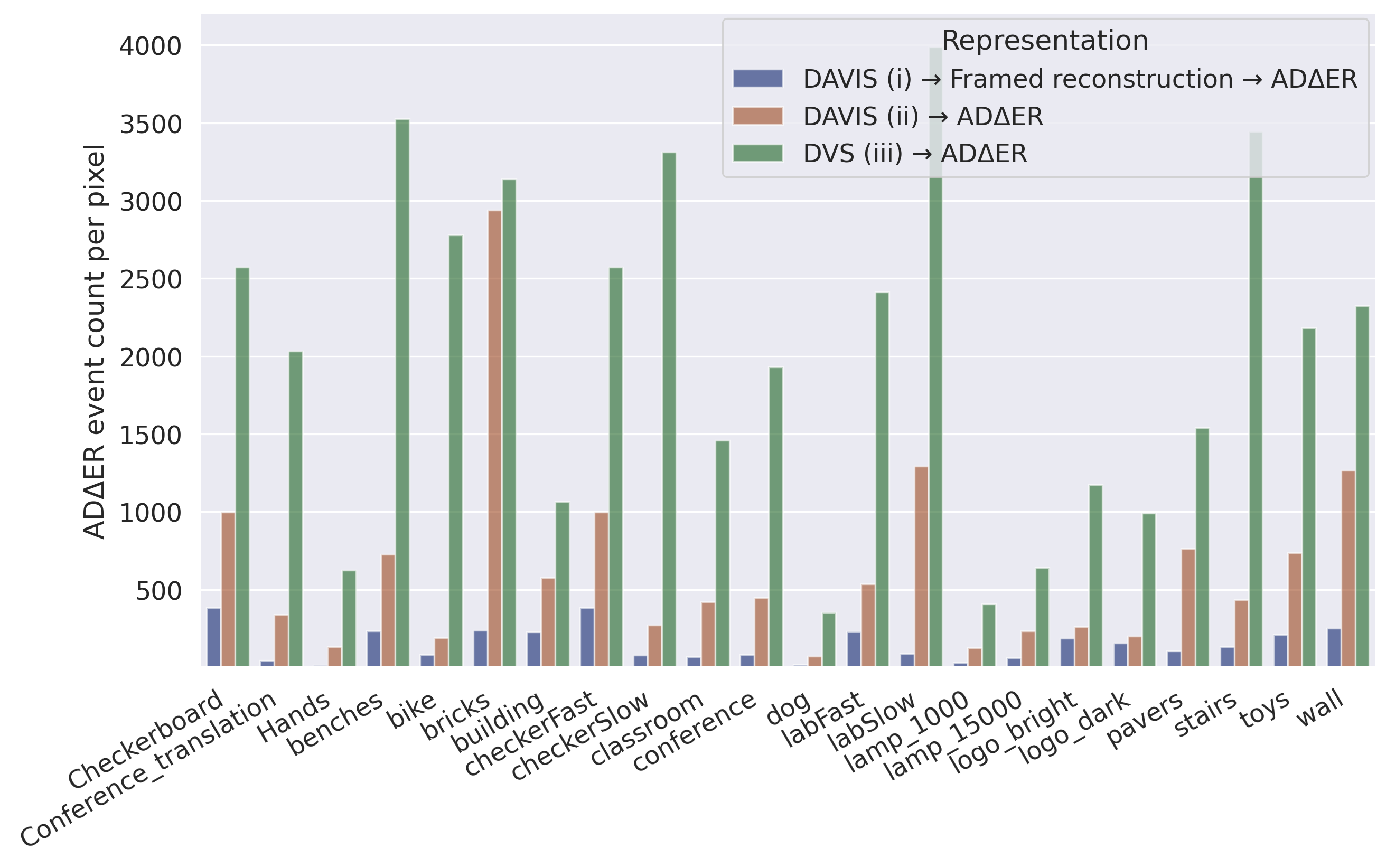}
        \label{fig:event_source_adder_rates}
        }
    \caption{Quantitative results of \eventformat{} on event camera sources. 
    (a) Effect of $\Delta t_{ref}$ on the latency of \eventformat{} transcoding. (b) Proportion of recovered DVS events after \eventformat{} transcoding, as compared to the source \texttt{.aedat4} DVS event rates. (c) \eventformat{} event rates for each transcode mode and each video of our data set, where $\Delta t_{ref} = \Delta t_{s}/500$ and $M=40$. Mode (iii) yields dramatically higher \eventformat{} rates, despite not encompassing absolute intensities in slowly-changing or static regions of the scene. }
    \label{fig:event_eval}
\end{figure*}

\subsubsection{Mode (ii): Deblurred APS frame + DVS events $\rightarrow$ \eventformat}

We may also choose to use EDI to simply deblur each APS frame, and input the DVS events occurring outside that frame's exposure time individually in our \eventformat{} transcoder. Since DVS events express log intensity relative to a previous latent value, we maintain a matrix of the most recent log intensity for each pixel, by which we calculate the intensity represented by the incoming DVS event. When we ingest our deblurred frame, we scale the frame intensities $I$ to the range $L \in [0,1]$, and set the latent log intensity for each pixel by $\widetilde{L} = \ln{(1 + L)}$. We interpret a DVS event as specifying the exact moment an intensity \textit{changes}. Suppose that, for a given pixel, we ingest a deblurred frame intensity of $I_0 = 20$ that spans time $[t'-\Delta t_{ref}, t']$, where the frame has 8-bit values and $\Delta t_{ref} = 1000$. Then, we have latent value $\widetilde{L_0} = 0.0755$. Suppose we have a sequence of DVS events $\langle p, t\rangle$ occurring after time $t'$, $\{E_1 = \langle 0, t' + 500\rangle, E_2 = \langle 1, t'+ 800\rangle, E_3 = \langle 1, t' + 1200\rangle\}$. When we encounter $E_1$, we first \textit{repeat} the previous integration to fill the time elapsed. That is, we integrate $\frac{L_0 \cdot 500}{\Delta t_{ref}} = 10$ intensity units over 500 ticks. 

Then, we simply set the new latent intensity as follows. Supposing $\theta = 0.15$, our latent value $L_1$ becomes $L_1 = \exp{(\widetilde{L_0} - 0.15)} - 1.0 = -0.0718$. Since we cannot integrate negative intensities, however, we must clamp $L_1$ and $\widetilde{L_1}$ to $0.0$. At this stage, we do not know how long the pixel maintains this intensity, so we do not immediately integrate an additional value. Rather, when we encounter $E_2$, we integrate $\frac{L_1 \cdot 255 \cdot 300}{\Delta t_{ref}} = 0$ intensity units over 300 ticks. As before, we set the new latent log value $\widetilde{L_2} = \widetilde{L_1} + 0.15 = 0.15$, such that $L_2 = \exp{(\widetilde{L_2})} - 1.0 = 0.1618$. When we encounter $E_3$, then, we integrate $\frac{L_2 \cdot 255 \cdot 400}{\Delta t_{ref}} = 41.27$ intensity units over 400 ticks.

Although the clamping mechanism causes a worse visual appearance than mode (i) (\cref{sec:mode_1}), as shown in the white and black pixels of moving edges in \cref{fig:davis_matrix}, this method preserves the temporal resolution of the source events. By contrast, intensity timings in mode (i) are quantized to the EDI reconstruction frame rate.

\subsubsection{Mode (iii): DVS events $\rightarrow$ \eventformat}\label{sec:mode_3}
Finally, we can choose to ignore the APS data, or use a DVS sensor alone, and integrate the DVS events directly. In this case, our EDI program functions solely as the camera driver or event file decoder. At sub-second intervals, we reset the latent log image $\widetilde{L}$ in the \eventformat{} transcoder to $\ln(0.5)$, representing a mid-level intensity. Then, we calculate the absolute intensity of the following DVS events relative to this mid-level intensity. As we do not deblur APS frames in this mode, we cannot find the optimal $\theta$ value by which to interpret these DVS events. Therefore, we use a constant $\theta =0.15$ to calculate the changes in log intensity.
As shown in \cref{fig:davis_pipeline}, we only gather intensity information for pixels that change greatly enough to trigger DVS events, and unchanging pixels will carry a mid-level gray value.

\subsection{Evaluation}
As our chief aim was to enable the development of fast, practical event-based systems, we needed to test our method on a real-world source representation. However, as discussed in \cref{sec:event_representations}, most event vision data sets use an intermediate representation that is slower to ingest. Thus, we employ the DVSNOISE20 data set, which provides raw \texttt{.aedat4} files in the same format that the DAVIS 346 camera produces. This data set provides 3 recordings for each of 16 scenes, though we used only one recording per scene in our evaluation. We also add to the data set 6 of our own recordings with varied APS exposure lengths and more extreme lighting conditions. 

 \begin{figure*}[t]
    \centering
    \subfloat[ ]{\includegraphics[width=0.405\linewidth]{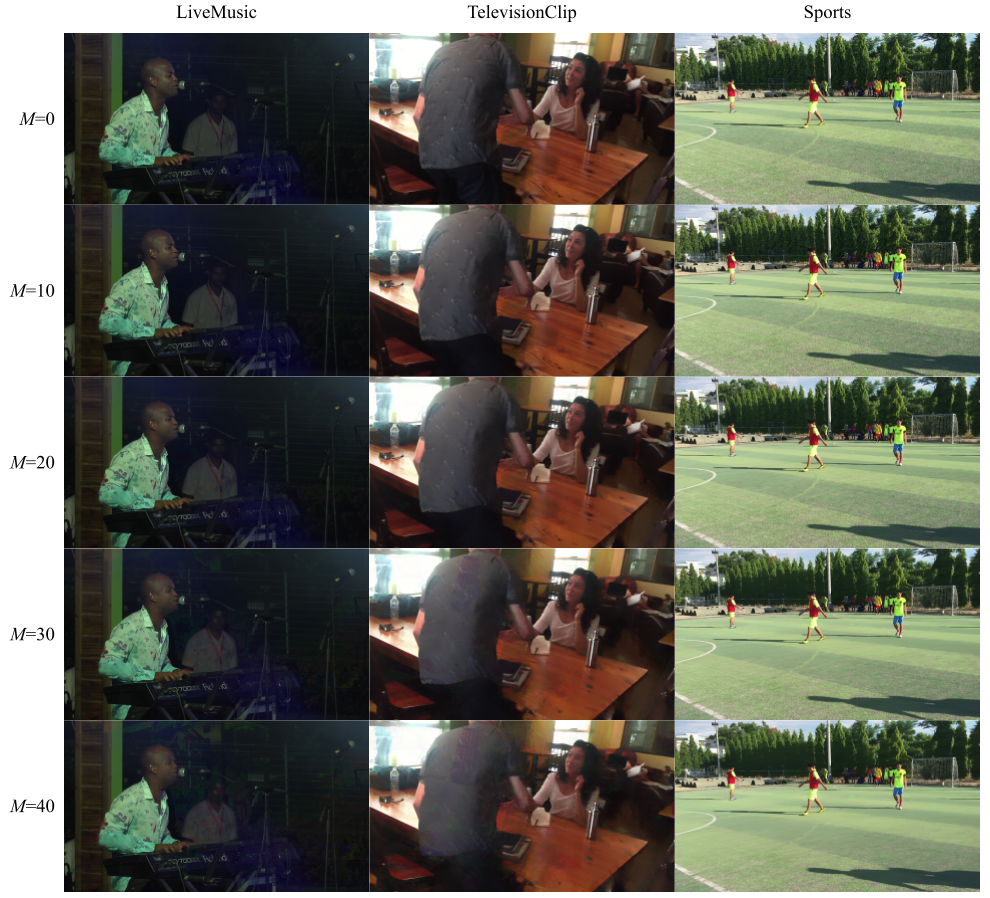}%
    \label{fig:framed_matrix}}
    \subfloat[ ]{\includegraphics[width=0.475\linewidth]{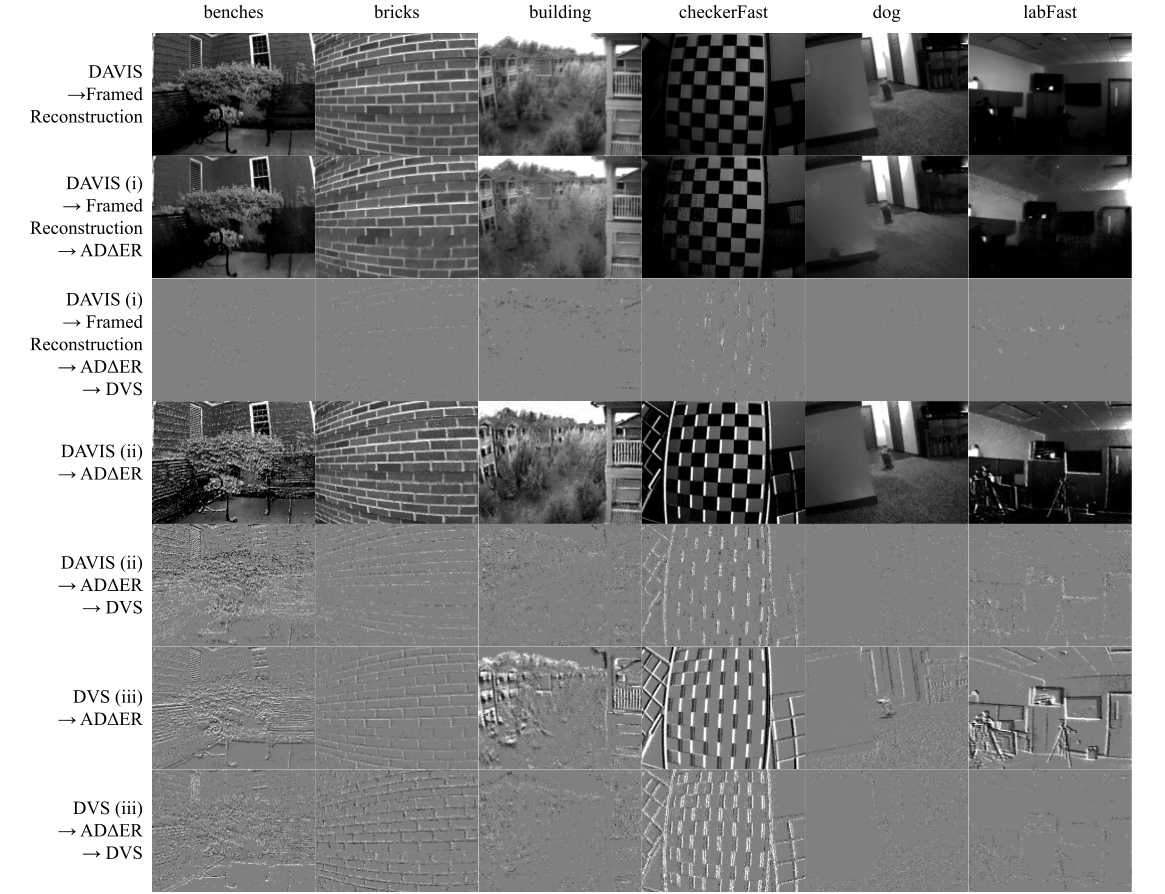}%
    \label{fig:davis_matrix}}
    \caption{Qualitative results of the \eventformat{} transcoder. (a) Instantaneous samples of framed videos transcoded to \eventformat{} at various $M$. High $M$ yields more smearing and ghosting artifacts in low-contrast regions. (b) Event videos transcoded to framed video (first row) and to \eventformat{} under our three transcode modes (rows 2, 4, and 6) with $\Delta t_{ref} = \Delta t_{s}/500$ and $M = 40$. We transcode our \eventformat{} representations back to DVS (rows 3, 5, and 7) and see that modes (ii) and (iii) preserve much more of the temporal contrast detail of the DVS streams.}
    \label{fig:matrices}
\end{figure*}

We transcoded each DAVIS video to \eventformat{} using mode (ii) with \eventformat{} threshold parameter $M \in \{0, 10, 20, 30, 40\}$, $\Delta t_{ref} \in \{100, 1000\}$, $\Delta t_{s} = 1\times 10^6$, and $\Delta t_{max} = \Delta t_{s} \cdot 4 = 4\times 10^6$. At $M=10$ for both choices of $\Delta t_{ref}$, we observed a median 49-50\% decrease in the \eventformat{} event rate compared to $M=0$. At $M=40$, we observed a median 66\% decrease in \eventformat{} event rate compared to $M=0$, demonstrating the utility of our temporal average scheme.


Secondly, we transcoded each DAVIS video to \eventformat{} under modes (i) and (ii) with a fixed $M=40$, and $\Delta t_{ref} \in \{\Delta t_{s}/1000,$ $\Delta t_{s}/500,$ $ \Delta t_{s}/250,$ $\Delta t_{s}/100,$ $\Delta t_{s}/75,$ $\Delta t_{s}/50\}$ ticks. Additionally, we ran direct framed reconstructions of DAVIS in EDI and we transcoded each video once with mode (iii). We recorded the maximum latency between reading a packet from the \texttt{.aedat4} source video in our EDI program and completing the transcode of that packet's data. In \cref{fig:davis_allmodes_latency}, we observe that both framed reconstruction and transcode mode (i) incur dramatically more latency as the effective frame rate increases, which corroborates our findings in \cref{sec:framed_evaluation}. By contrast, mode (ii) incurs relatively constant $2.1$ second latency across all values of $\Delta t_{ref}$, yet has less variance as $\Delta t_{ref}$ increases. The speed of mode (iii) does not depend on $\Delta t_{ref}$, and we see that we can achieve a median transcode latency of only 0.4 seconds. These results support our argument that it is computationally expensive to employ a high-rate, framed intensity representation for event data. While such methods unavoidably quantize the temporal components of the event sources, modes (ii) and (iii) of our scheme can transcode the DAVIS/DVS event data to an intensity representation, \eventformat{}, with a temporal resolution matching that of the source.

We see direct evidence for this claim in \cref{fig:back_to_dvs_event_rates}, which shows the proportion of DVS events reconstructed from the \eventformat{} representations after transcoding with $M = 40$ and $\Delta t_{ref} = \Delta t_{s}/500$ ticks, compared to the number of DVS events in the \textit{source} DAVIS videos. We can quickly reconstruct a DVS stream from \eventformat{}, since the intensity changes to trigger DVS events will occur only at the temporal boundaries of our sparse \eventformat{} events. However, we must assume a constant $\theta$, as in transcode mode (iii) (\cref{sec:mode_3}). Mode (i), which quantizes events into fixed-duration frames, can recover only a small fraction of the DVS events in the source. On the contrary, modes (ii) and (iii) have high proportions of recoverable DVS events, even with a high $M$. \cref{fig:event_source_adder_rates} illustrates that transcode mode (ii) generally necessitates far fewer \eventformat{} events than mode (iii) with these same settings. Thus, our \eventformat{} codec with mode (ii) can simultaneously utilize the APS frames to compress the DVS stream and utilize the DVS stream to deblur the APS frames, while fusing both streams into a unified, asynchronous intensity representation.

\cref{fig:davis_matrix} shows qualitative results on six scenes from our data set\footnote{We provide the framed reconstructions of these videos as supplementary material \href{https://drive.google.com/drive/folders/1VXAMPrNnm0qRDqyz0pqDV_dJTNh-h_hm?usp=sharing}{here}.}. The \eventformat{} transcodes shown use $M=40$ and $\Delta t_{ref} = \Delta t_{s}/500$, as in our latency experiments. Since mode (i) of our event transcoder simply uses our framed source transcoder (\cref{sec:framed_transcoder_details}) after reconstructing a frame sequence with EDI, we note that the artifacts in the \eventformat{} representation here stem from artifacts in EDI's \textit{framed reconstruction}, or from the temporal averaging in \eventformat{}. In particular, we point to the \texttt{building} images, which show the limitations of EDI in deblurring APS frames under fast motion and long exposures. We see that we recover few DVS events from mode (i), due to its temporal quantization of the source events. However, mode (ii) exhibits a similar quality of recovered DVS events compared to mode (iii), despite also conveying the absolute intensities for pixels without DVS source events.



\section{Future Work}
\subsection{Extending this Paper}
We note the difficulty in empirically evaluating the quality of our transcoded representation for DAVIS/DVS sources. Since APS frames are blurred, they are not reliable references for framed quality metrics (e.g., PSNR, SSIM, and VMAF). Furthermore, existing metrics for DVS stream quality focus on noise filtering tasks and can only process a few hundred events per second \cite{spike_metric}, while real-world DVS sequences encompass several millions of events per second. Therefore, when an end-to-end system is in place, we will evaluate application performance as a metric for the representational quality of streams transcoded from event-based sources. Finally, we will also work to expand our framework with support for additional event camera data types, such as those from Prophesee \cite{prophesee_detection}.

\subsection{Bigger Picture}
Our video acquisition and representation pipeline is a necessary first step toward the development of a real-time, source-agnostic event video ecosystem. The next steps lie in the compression, streaming, and applications of \eventformat{} data.

As we have previously noted, \eventformat{} is amenable to lossy compression, yet incompatible with traditional video codecs. Therefore, it requires novel techniques for inducing loss according to metrics of data distortion, connection bandwidth, and application performance. \eventformat's sparse representation naturally produces low data rates where pixels are the least dynamic, but compression techniques must discern how to induce intensity accuracy loss without causing temporal decoherence between independent pixel streams (i.e., from events' $\Delta t$ values losing different levels of precision).

Finally, connected client applications can take a two-pronged approach to \eventformat{} data: they may be bespoke event-based applications ingesting \eventformat{} streams directly, or they may be classical frame-based applications seeking to benefit from our sparse representation and high temporal precision. In the latter case, the client can utilize our open-source library to reconstruct framed images from received \eventformat{} data, though the maximum incurred reconstruction latency will scale linearly with the choice of \dtm{} in the case of real-time streaming. Computational overhead will also limit the spatiotemporal resolution that may be processed in real time, though robust compression schemes can help alleviate this barrier.

\section{Conclusion}
This paper proposes the \eventformat{} video representation for both framed and event sources. We show that, through temporal averaging of similar intensities, we can increase the intensity precision and greatly reduce the number of samples per pixel compared to grid-based representations. We demonstrate extremely low-latency DAVIS fusion, and we argue that our method provides the computational efficiency and temporal granularity necessary for building real-time intensity-based applications for event cameras.

\bibliographystyle{ACM-Reference-Format}
\bibliography{paper}










\end{document}